\documentclass{elsarticle}[review]
\usepackage{latexsym}
\usepackage{graphicx}
\usepackage{epsfig}
\usepackage{url}
\usepackage{amsmath,amssymb}
\numberwithin{equation}{section}

\graphicspath{ {./figures/} }
\usepackage{hyperref}
\usepackage{float}
\usepackage{verbatim}
\usepackage{color}
\usepackage{amsmath}
\usepackage{amsfonts}

\usepackage{booktabs}
\usepackage{algorithm}
\usepackage{algorithmic}
\usepackage{soul}
\usepackage{multirow}
\usepackage{color}
\usepackage{subcaption}
\usepackage[draft]{todonotes}   
\usepackage{enumerate}
\usepackage{wasysym}

\usepackage[utf8]{inputenc}
\usepackage[table]{xcolor}
\usepackage{hhline}
\usepackage{booktabs}
\usepackage{xspace}
\usepackage{caption}

\newcommand{\name}{TIGRAG}%
\newcommand{\link}{https://github.com/FedericaParlapiano/TIGRAG}
\begin{document}
\begin{frontmatter}

\title{Efficient Retrieval-Augmented Generation via Token Co-occurrence Graphs}

\author[1]{Gianluca Bonifazi}
\ead{g.bonifazi@univpm.it}
\author[1]{Christopher Buratti}
\ead{c.buratti@pm.univpm.it}
\author[1]{Michele Marchetti}
\ead{michele.marchetti@univpm.it}
\author[1]{Federica Parlapiano}
\ead{f.parlapiano@pm.univpm.it}
\author[1]{Giulia Quaglieri}
\ead{g.quaglieri@pm.univpm.it}
\author[2]{Davide Traini}
\ead{davide.traini@unimore.it}
\author[1]{Domenico Ursino}
\ead{d.ursino@univpm.it}
\author[1]{Luca Virgili\corref{cor1}}
\ead{luca.virgili@univpm.it}

\cortext[cor1]{Corresponding author}
\fntext[fn1]{This is the first author footnote.}
\fntext[fn2]{Another author footnote, this is a very long
footnote and it should be a really long footnote. But this
footnote is not yet sufficiently long enough to make two
lines of footnote text.}
\fntext[fn3]{Yet another author footnote.}
\affiliation[1]{organization={Università Politecnica delle Marche},
city={Ancona},
    country={Italy}}
\affiliation[2]{organization={Università di Modena e Reggio Emilia},
city={Modena},
    country={Italy}}

\date{}


\begin{abstract}
Retrieval-Augmented Generation (RAG) mitigates hallucinations in Large Language Models (LLMs) by grounding the generation process on external knowledge. However, standard RAG approaches struggle with multi-hop reasoning. While recent graph-based RAG methods improve the retrieval of interconnected chunks, they often rely on computationally expensive and error-prone LLM-based extraction pipelines. To address these issues, we propose \name{} (Token-Induced GraphRAG), an efficient graph-augmented RAG framework based on a token co-occurrence Knowledge Graph. \name{} directly models topological relationships between tokens using sliding-window co-occurrence statistics, thus enabling scalable graph construction. During inference, it combines graph-based semantic expansion and neural reranking to retrieve interconnected evidence for multi-hop reasoning. Specifically, it introduces an iterative entity-driven retrieval strategy that progressively expands the query using bridging entities extracted from previously retrieved contexts. We evaluated \name{} on three widely adopted multi-hop Question Answering (QA) benchmarks. Experimental results demonstrated that our framework consistently outperforms dense retrieval and graph-based RAG methods in both retrieval and downstream QA tasks, while substantially reducing indexing time, inference latency, and prompt footprint.
\end{abstract}

\end{frontmatter}

\section{Introduction}
\label{sec:Introduction}


As Large Language Models (LLMs) become more widespread, the research community is paying increasing attention to their inherent vulnerabilities, particularly the hallucination phenomenon \cite{PeDrBo24, Yao*23}. Retrieval-Augmented Generation (RAG) has emerged as a well-established approach to mitigate hallucinations by anchoring LLM outputs in external, factual knowledge \cite{Gao*23-2, Peng*25, Cheng*26}. In a standard RAG pipeline, a user query is encoded as a vector, which is then employed to retrieve relevant text chunks from a large document corpus through a similarity search \cite{Malik*26, Jeong*25}. This retrieved information is then used to provide the LLM with contextual evidence during generation \cite{Kuratov*24, Liu*24-6, YuXuAk24}. However, traditional vector-based RAG systems are ineffective because they treat documents as largely disconnected units, which limits their ability to reason across multiple pieces of information. Consequently, they often struggle with multi-hop queries, which require synthesizing dispersed evidence across independent chunks \cite{AmGoAk25,TaYa24,Ma*25, ZhLi26}. 

To address this issue, Graph-augmented RAG (GraphRAG) approaches rely on structured representations, often formalized as Knowledge Graphs (KGs). In KGs,  nodes and edges capture textual elements and their complex relationships, which provides a topological space to map and navigate knowledge \cite{Edge*24, Guo*25-2}. These approaches range from injecting graph-derived knowledge (such as triplets or community summaries) directly into the context \cite{Edge*24}, to using the topological structure as a routing mechanism to retrieve the most relevant raw text chunks \cite{Gutierrez*24}. Due to their interconnected structure, GraphRAG approaches can inherently address the complex reasoning paths required by multi-hop Question Answering (QA) \cite{SaTuSa24, Jin*24}. Nevertheless, two critical challenges hinder their broader adoption. First, offline KG construction introduces significant computational overhead and latency \cite{Min*25, Choubey*26}. Second, traditional GraphRAG pipelines \cite{Edge*24, Guo*25-2} rely heavily on LLM-based entity and relationship extraction, making them susceptible to hallucinated, noisy, or missing connections. These graph construction errors can propagate through the retrieval pipeline and ultimately cause cascading failures in downstream reasoning and generation \cite{GhCr25}.

This paper aims to address these computational overheads and extraction-based vulnerabilities by proposing \name{} (Token-Induced GraphRAG), a highly efficient graph-augmented RAG. Unlike traditional GraphRAG systems, \name{} statistically constructs its topological foundation by extracting tokens from the corpus and linking them based on co-occurrences within a sliding window. \name{}'s architecture consists of four core components: {\em (i)} an offline graph extraction stage that efficiently builds the topological KG, {\em (ii)} a semantic expansion and candidate retrieval module powered by Personalized PageRank (PPR) \cite{Haveliwala02}, {\em (iii)} a dynamic graph-weighted scoring mechanism for contextual chunk evaluation, and {\em (iv)} a final neural reranking module that distills the top chunks based on semantic alignment with the query.

We evaluated \name{} using the standard protocols outlined in the literature \cite{Gutierrez*24, Gutierrez*25} on three multi-hop QA benchmarks: HotpotQA \cite{Yang*18}, 2WikiMultiHopQA \cite{Ho*20}, and MuSiQue \cite{Trivedi*22}. Our empirical results demonstrate that \name{} not only outperforms existing RAG baselines across all datasets but also yields substantial efficiency improvements. Indeed, it significantly reduces indexing and retrieval latency, as well as overall token consumption.

This paper is organized as follows: Section~\ref{sec:Related-Work} discusses related work. Section~\ref{sec:Proposed-Approach} details the \name{}'s architecture and its core algorithms. Section~\ref{sec:Experimental-Campaign} describes the experimental campaign performed to evaluate \name{}. Section~\ref{sec:Discussion} presents a discussion on our framework. Finally, Section~\ref{sec:Conclusion} draws our conclusions and outlines some promising directions for future research.

\section{Related Work}
\label{sec:Related-Work}

This section illustrates approaches related to \name{}. Specifically, Subsection \ref{sub:Standard-RAG} presents standard RAG approaches. Subsection \ref{sub:Graph-Augmented} describes Structure-Augmented Retrieval Approaches. Finally, Subsection \ref{sub:Network-Based-Retrieval} discusses the Network-Based Retrieval and Multi-Hop Reasoning paradigms.

\subsection{Standard Retrieval-Augmented Generation}
\label{sub:Standard-RAG}

RAG has emerged as a key paradigm for limiting LLM hallucinations. It anchors LLM outputs in reliable and up-to-date external knowledge sources, thereby improving factual accuracy and response quality \cite{Fan*24}.

Within this established paradigm, traditional RAG frameworks typically use lexical searches \cite{Ram*23} or semantic similarities \cite{Karpukhin*20} to retrieve the necessary context. These approaches encode text chunks as continuous vectors and apply similarity metrics, such as cosine similarity, to effectively retrieve documents that share semantic overlap with the input query. However, standard RAG approaches treat documents as isolated units. Consequently, they have significant difficulty with complex, multi-hop queries that require synthesizing information across disparate texts. In fact, in this case they are limited by the lexical and semantic gap between the initial question and the distant supporting facts \cite{AmGoAk25, TaYa24}.

Furthermore, standard dense retrieval methods often retrieve overly long contexts, resulting in severe information redundancy and substantial computational overhead. To address this issue, recent studies have explored context pruning techniques. For instance, AttentionRAG \cite{Fang*25} introduces an attention-guided compression mechanism to reduce token consumption and filter out irrelevant information from the retrieved passages. While context compression offers a partial remedy, it often struggles to balance compression rates with the risk of information loss. Therefore, a more structural solution is needed to accurately identify interconnected evidence instead of just compressing flat text.

\subsection{Structure-Augmented Retrieval Approaches}
\label{sub:Graph-Augmented}

To overcome the limitations of flat vector spaces, recent approaches have integrated structural dependencies into the RAG pipeline. In this context, GraphRAG has emerged as a promising paradigm that leverages graph-based indexing and retrieval to exploit relationships among entities. This allows for more precise and context-aware reasoning over complex knowledge bases \cite{Peng*25}.

Several specific architectures have been proposed based on this paradigm. For instance, RAPTOR \cite{Sarthi*24} organizes textual knowledge into recursive tree structures through iterative abstractive processing, which facilitates retrieval at various levels of granularity. Similarly, TreeRAG \cite{Tao*25} addresses the fragmentation typical of long documents by employing tree-based chunking and bidirectional traversal strategies. This approach strictly preserves hierarchical dependencies during retrieval. 

Concurrently, graph-based systems have gained significant attention for their ability to improve information aggregation through hierarchical indexing and summarization \cite{Sarmah*24}. GraphRAG \cite{Edge*24} leverages KGs to generate hierarchical community-level summaries, while LightRAG \cite{Guo*25-2} integrates graph topologies with dual-level vector representations to efficiently retrieve entities and relationships. Approaches like $KG^2RAG$~\cite{Zhu*25-2} and HyperGraphRAG~\cite{Luo*25-3} further exploit structured inter-chunk relationships by using KGs and hypergraph representations, respectively, to enrich the retrieved information. Furthermore, GFM-RAG \cite{Luo*25-2} introduces Graph Foundation Models to directly augment the generative process with deep structural awareness. ToG \cite{Sun*24-2} seamlessly integrates information retrieval from both structured KGs and unstructured text, employing a tightly coupled reasoning mechanism to ground the generative process. KGP \cite{Wang*24-5} explicitly targets Multi-Document QA (MD-QA) by constructing a passage-level KG and deploying an LLM-based traversal agent. This dual mechanism regulates the transition space among interconnected documents while dynamically gathering relevant context, thereby balancing retrieval latency and generation quality.

Many of these approaches are effective, but they rely on computationally expensive LLM-driven extraction pipelines or extensive structural pre-training to construct their underlying graphs \cite{Xiao*25}. Furthermore, recent studies have shown that implementations like GraphRAG encounter significant scalability issues in real-world, large-scale deployments. In these scenarios, the computational complexity of graph traversal algorithms and retrieval latency increase exponentially with the size of the KG \cite{Peng*25}. 

\name{} builds a highly traversable KG directly from normalized text tokens and their local co-occurrences. This streamlined approach bypasses the heavy computational cost of generative graph construction, ensuring highly efficient, low-latency retrieval operations while preserving the intrinsic structural integrity of the corpus.

\subsection{Network-Based Retrieval and Multi-Hop Reasoning}
\label{sub:Network-Based-Retrieval}

While many approaches focus on structuring the knowledge index for better summarization, another line of research emphasizes the direct use of graph topologies to retrieve interconnected contexts \cite{Ji*24,Wang*24-6}. For example, GRAG \cite{Hu*25} combines textual and topological information to pinpoint relevant subgraphs, whereas KG-RAG \cite{LiTo25} employs explicit question decomposition and multi-hop KG retrieval to enable deep reasoning. However, graph-based retrieval approaches often face a granularity tradeoff \cite{Wu*25}. In fact, fine-grained, entity-level graphs can capture detailed relationships but are computationally expensive and may lose contextual information. Conversely, coarse document-level graphs preserve context but struggle to model complex, multi-hop relationships. Recent methods, such as QCG-RAG \cite{Wu*25}, address this issue by building query-centric graphs to retrieve relevant chunks at different levels of granularity. Other approaches rely on Graph Neural Networks (GNNs) with query-aware attention mechanisms to dynamically identify and aggregate relevant information from KGs \cite{AgWaPu25}.

Rather than relying on explicit query decomposition or training computationally intensive GNN architectures, a parallel and highly effective strategy leverages network theory algorithms to organically navigate these knowledge structures. Approaches like HippoRAG \cite{Gutierrez*24} and its successor HippoRAG2 \cite{Gutierrez*25} have demonstrated the effectiveness of using PPR to emulate human cognitive indexing. This allows the retrieval signal to naturally spread across semantic nodes. 

While \name{} is inspired by these network-based principles, it introduces a token-level extraction phase, a dynamic candidate pooling strategy, and an iterative, entity driven multi-hop retrieval process. In particular, it extracts Named Entities from the first retrieval hop to expand the query and applies a neural reranking step. This allows it to retrieve interconnected evidence that traditional dense or single-hop graph-based methods often overlook.

\section{Proposed Approach}
\label{sec:Proposed-Approach}

This section outlines the technical features of \name{}. Specifically, Subsection \ref{sub:Motivations} explains the rationale behind our framework, whereas the following subsections describe the various steps that comprise its underlying approach.

\subsection{Motivations}
\label{sub:Motivations}

Standard RAG systems often struggle with multi-hop reasoning, whereas graph-based retrieval approaches frequently incur the cost of complex LLM-driven graph construction. To address these issues, we propose \name{}, a lightweight graph-augmented retrieval framework based on token co-occurrence graphs. Instead of relying on LLM-driven graph construction, \name{} builds a token-level topological representation directly from textual co-occurrences, enabling efficient and scalable graph creation. During retrieval, it combines graph-based semantic expansion with neural reranking to identify interconnected evidence for multi-hop reasoning. Then, an iterative, entity-driven retrieval strategy progressively augments the query using bridging entities extracted from previously retrieved contexts. \name{} substantially reduces prompt size, inference latency, and overall computational overhead by dynamically selecting compact and highly relevant contexts, outperforming existing GraphRAG approaches. 

\begin{figure}[ht!]
    \centering
    \includegraphics[width=\textwidth]{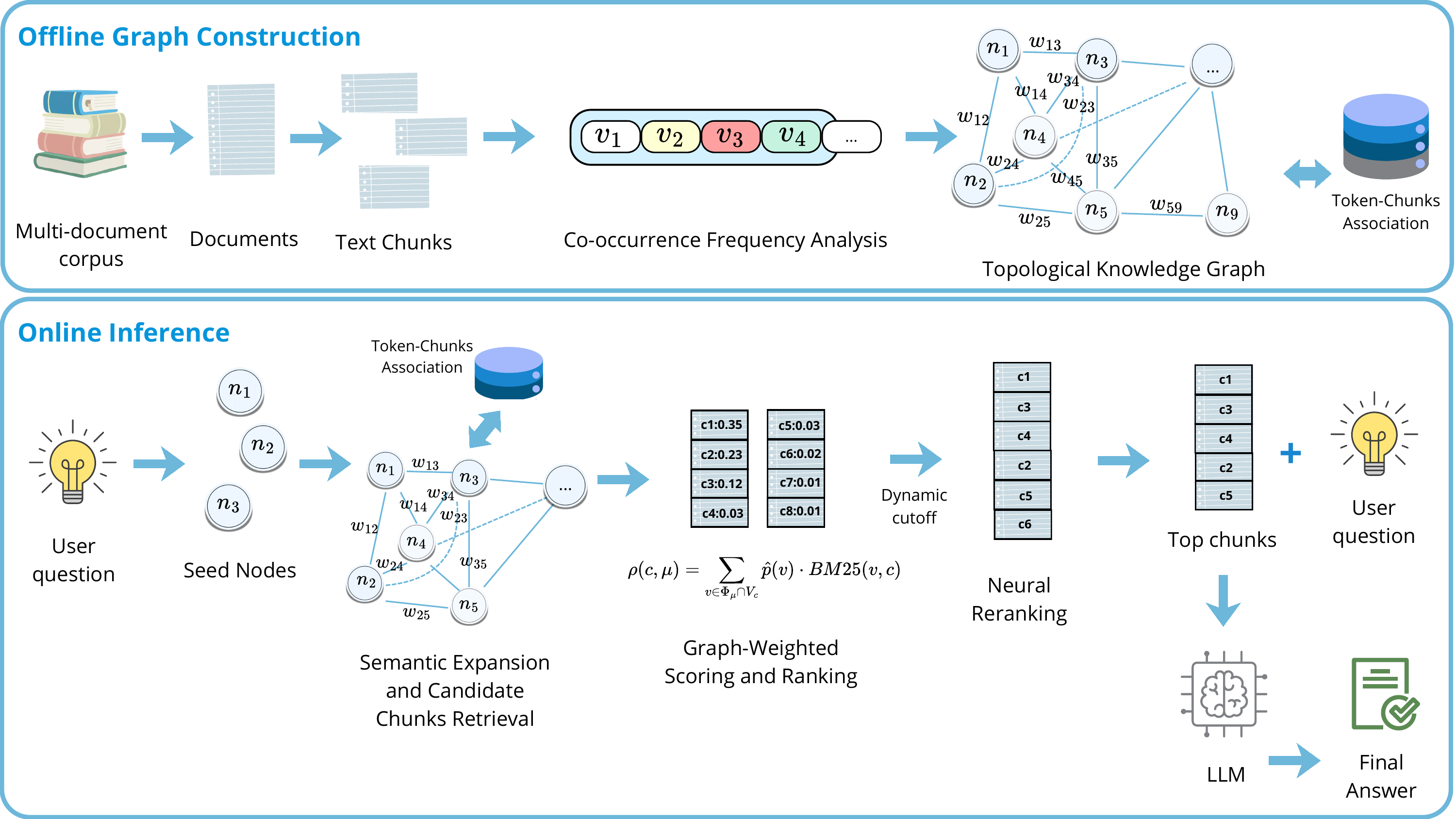}
    \caption{Workflow of \name{}, showing graph construction, semantic expansion, and iterative multi-hop retrieval}
    \label{fig:workflow}
\end{figure}

Figure~\ref{fig:workflow} illustrates the \name{}'s workflow. As shown in this figure, the \name{} architecture operates across two fundamental stages: {\em (i)} Offline Graph Construction, and {\em (ii)} Online Inference. During the offline phase, \name{} segments the raw corpus into overlapping chunks, from which token co-occurrence frequencies are used to construct the topological KG. At the same time, \name{} precomputes and indexes chunk-level dense embeddings and token-to-chunk provenance maps. At inference time, it tokenizes the incoming query to identify topological seed nodes. These seeds then drive a semantic expansion process that fetches an initial pool of candidate chunks. Next, \name{} applies a graph-weighted scoring mechanism to dynamically filter the candidates before passing them to a neural reranker for precise semantic alignment. Ultimately, it injects the top distilled chunks into the LLM's prompt to ground the final generation. This single-hop retrieval process is then repeated to enable multi-hop reasoning. 

\subsection{Pre-processing and Text Chunking}
\label{sub:Preprocessing-Text-Chunking}

Let $\mathcal{D} = \{d_1, \dots, d_f\}$ be the initial corpus of $f$ documents. Each document $d_i \in D$ is defined as a tuple $\langle t_i, b_i \rangle$, representing its title and its body, respectively. First, we apply a sentence tokenizer to segment the body $b_i$ into an ordered sequence of sentences $S_i = (\varrho_{i_1}, \dots, \varrho_{i_m})$. Afterwards, we perform a preprocessing pipeline to prepare the raw corpus for subsequent graph construction. Since TIGRAG relies on token co-occurrences, preserving local semantic coherence during chunking is essential to avoid noisy graph connections. 

Although recent attention-based segmentation strategies \cite{Guo*25-3} improve semantic preservation, they introduce additional computational overhead. Therefore, \name{} does not adopt these strategies but uses a lightweight, sentence-level, sliding-window strategy.

Let $s$ be the window size (i.e., number of sentences per chunk) and let $o$ be the overlap. We define the stride length as $l = s - o$. Finally, let $\lambda_i$ be the number of chunks of $d_i$. The $k$-th chunk of $d_i$, $1 \leq k \leq \lambda_i$, denoted as $c_{i_k}$, is formally defined as the subsequence:
\begin{equation}
\label{eq:c_i_k}
    c_{i_k} = (\varrho_{i_{(k - 1)l + 1}}, \dots,\, \varrho_{i_{(k - 1)l + s}})
\end{equation}

To facilitate the computation of semantic similarity during the retrieval phase, we map each chunk $c_{i_k}$ into a dense vector representation. Specifically, we use an embedding model $\mathcal{M}$ to compute the chunk-level embedding $\epsilon_{i_k} \in \mathbb{R}^\delta$:
\begin{equation}
    \epsilon_{i_k} = \mathcal{M}(c_{i_k})
\label{eq:epsilon_i_k}
\end{equation}

where $\delta$ is the embedding dimensionality. We store these dense representations and use them for cosine similarity-based retrieval (see below). This enables us to efficiently select the most relevant chunks for generation. 

After chunking, we tokenize each chunk $c_{i_k}$ through a function $\tau(\cdot)$ that receives a chunk and returns the set of its tokens. To reduce noise, we remove stopwords that could negatively affect the graph structure. Let $\mathcal{S}$ be the set of stopwords and let $\phi(\cdot)$ be a boolean validation function that takes a token as input and enforces minimum length and numerical constraints (i.e., it preserves years and large numerical values while discarding punctuation and tokens shorter than three characters). The final set $V_{i_k}$ of valid tokens extracted from $c_{i_k}$ is defined as:
\begin{equation}
\label{eq:V_i_k}
    V_{i_k} = \{ v \in \tau(c_{i_k}) \mid v \notin \mathcal{S} \land \phi(v) \}
\end{equation}

The tokens of ${V_{i_k}}$ form the basis for constructing the KG supporting TIGRAG, as explained in the next subsections.

\subsection{Knowledge Graph Construction}
\label{sub:Knowledge-Graph-Construction}

After extracting the valid token sets $V_{i_k}$ for each chunk $c_{i_k}$, we construct a weighted, undirected KG:

\begin{equation}
\label{eq:G}
\mathcal{G} = \langle N, E \rangle
\end{equation}

$N$ is the set of nodes in ${\cal G}$. There is a node $n_p \in N$ for each token $v_p \in V$. $V$ is defined as the union of all valid tokens extracted across the entire corpus. Formally speaking:

\begin{equation}
\label{eq:V}
    V = \bigcup_{i=1..f, k=1..\lambda_i} V_{i_k}
\end{equation}

Since each node $n_p \in N$ corresponds to exactly one token $v_p \in V$, we use the terms ``node'' and ``token'', as well as the symbols $n_p$ and $v_p$, interchangeably. 

$E$ is the set of edges in ${\cal G}$. It is based on token co-occurrences in such a way as to establish semantic relationships between extracted concepts. Specifically, for each processed chunk $c_{i_k}$, let $V_{i_k}$ represent the ordered sequence of its valid tokens. We traverse this sequence using a sliding window of size $w$ and add an edge $(n_p,n_q,\omega_{pq})$ to $E$ if the tokens $v_p$ and $v_q$ co-occur at least once within this window. The weight $\omega_{pq}$ is equal to the cumulative frequency of the co-occurrences of $v_p$ and $v_q$ across all sliding windows in the corpus. It reflects the strengths of the semantic association between $v_p$ and $v_q$.

\subsection{Query Pre-processing and Semantic Expansion}
\label{sub:Query-Preprocessing}

During inference, an incoming natural language query $\mu$ undergoes the same preprocessing and tokenization pipeline used for graph construction. Let $V_\mu$ denote the set of valid tokens extracted from $\mu$, and let $V_{seed} = V_\mu \cap V$ denote the subset of the tokens of $\mu$ appearing in ${\cal G}$. We use the nodes corresponding to $V_{seed}$ to perform a PPR process over $\mathcal{G}$. This allows us to explore the topological neighborhood of $\mu$ and identify semantically related concepts.

Specifically, we define a personalization vector $\mathbf{v} \in \mathbb{R}^{|V|}$, where the restart probability is uniformly distributed across the nodes of $V_{seed}$ and set to zero otherwise. In particular, the component $\mathbf{v}_p$ of $\mathbf{v}$, corresponding to the node $v_p \in V$, is defined as follows:

\begin{equation}
\label{eq:v_p}
\mathbf{v}_p = \begin{cases} \frac{1}{|V_{seed}|} & \text{if } v_p \in V_{seed} \\ 0 & \text{otherwise} \end{cases}
\end{equation}

We then execute PPR on $\mathcal{G}$ using $\mathbf{v}$ as the personalization vector. Upon convergence, we sort the nodes of ${\cal G}$ according to their PageRank scores, and select the top-$\gamma$ nodes to form the expanded token set $\Phi_\mu$ corresponding to $\mu$. Finally, we $L_1$-normalize the corresponding PPR scores to obtain a discrete probability score $\hat{p}(v)$ for each token $v \in \Phi_\mu$.

\subsection{Candidate Retrieval and Graph-Weighted Scoring}
\label{sub:Candidate-Retrieval}

After the expansion step, we retrieve a set of candidate text chunks. A chunk $c$ is considered a candidate if it contains at least one token from the union of the original query tokens and the expanded set of tokens, i.e., from $V_\mu \cup\ \Phi_\mu$. 

We introduce a graph-weighted scoring to rank these candidate chunks. In particular, we combine the probabilistic retrieval capabilities of BM25 \cite{SvBu09} with the semantic information provided by the PPR expansion. The relevance score $\rho(c,\mu)$ for a candidate chunk $c$, given the query $\mu$, is defined as the sum of the BM25 contributions of the tokens of $\Phi_\mu$ present in $c$, weighted by their PPR:

\begin{equation}
\label{eq:sigma}
\rho(c,\mu) = \sum_{v \in \Phi_\mu \cap V_c} \hat{p}(v) \cdot BM25(v,c)
\end{equation}

Here, $V_c$ is the set of tokens within $c$, $BM25(v,c)$ represents the standard BM25 score contribution of the token $v$ of $c$, and $\hat{p}(v)$ acts as a topological modulating factor derived from the PPR distribution. This formulation ensures that candidate chunks are ranked not merely by superficial lexical overlap but also by the structural relevance of the underlying concepts.

\subsection{Dynamic Candidate Pooling and Reranking}
\label{sub:Dynamic-Candidate-Pooling}

Although graph-weighted scoring effectively retrieves a broad set of contextually relevant chunks, relying strictly on a fixed cutoff can introduce noise. To dynamically filter the candidate set, we apply a cumulative score thresholding mechanism. 

Let $\mathcal{C}_\mu$ be the list of retrieved candidate chunks sorted in descending order by their hybrid score $\rho(\cdot,\cdot)$. First, we calculate the total retrieval score mass $\rho_{\mu_{total}}$ as the sum of all the scores of all chunks in ${\cal C}_\mu$:

\begin{equation}
\label{eq:sigma_total}
\rho_{\mu_{total}} = \sum_{c_k \in \mathcal{C}_\mu} \rho(c_k,\mu)
\end{equation}

Then, we construct a refined candidate pool $\mathcal{P}_\mu$ of chunks by selecting the top $\alpha$ chunks such that their cumulative score reaches a predefined fractional threshold $th_{\rho}$:
\begin{equation}
\label{eq:alpha}
     \alpha = \min_{m=1 .. |\mathcal{C}_\mu|} \left( \frac{\sum_{k=1}^m \rho(c_k,\mu)}{\rho_{\mu_{total}}} \ge th_{\rho} \right) 
\end{equation}

This dynamic cutoff ensures that TIGRAG retains only the most significant chunks, automatically adjusting the size of the pool based on the model's confidence level.

Finally, to ensure the maximum semantic fidelity to the query $\mu$, we perform a neural reranking on the dynamic pool $\mathcal{P}_\mu$. To this end, we use the embedding model $\mathcal{M}$ to project the natural language query $\mu$ into the same continuous vector space as the precomputed chunk embeddings, yielding $\epsilon_\mu = \mathcal{M}(\mu) \in \mathbb{R}^\delta$. For each candidate chunk $c \in \mathcal{P}_\mu$ with an embedding $\epsilon_c$, we determine the final ranking by computing the cosine similarity $\sigma_{c \mu}$ between the vectors $\epsilon_c$ and $\epsilon_\mu$. After constructing $\mathcal{P}_\mu$, we reorder its chunks $c \in \mathcal{P}_\mu$ in descending order according to the cosine similarity $\sigma_{c \mu}$. This yields the definitive ranking for the initial retrieval phase that returns the results of the one-hop query. We call $\hat{\mathcal{P}}_\mu$ the ordered list of chunks thus obtained.

\subsection{Multi-Hop Reasoning and Entity-Driven Query Expansion}
\label{sub:Multi-Hop-Reasoning}

To address complex, multi-hop queries, i.e., queries that may lack the explicit vocabulary required to retrieve all supporting evidence, we introduce an iterative, entity-driven retrieval mechanism. At each hop $j$, let $\mathcal{H}_\mu^j$ denote the list of the top $\iota$ retrieved chunks sorted by the similarity degrees between their embeddings and the embedding of $\mu$. Clearly, $\mathcal{H}_\mu^1$ coincides with the top $\iota$ chunks of  $\hat{\mathcal{P}}_\mu$. Finally, we denote by $\sigma^j_{c \mu}$ the cosine similarity between the embeddings $\epsilon_c$ and $\epsilon_\mu$ at the hop $j$. Clearly, $\sigma^1_{c \mu}$ coincides with $\sigma_{c \mu}$. We process the chunks of $\mathcal{H}_\mu^j$ through a Named Entity Recognition (NER) pipeline to extract the set of salient entities:

\begin{equation}
\label{eq:N_j}
    {\cal N}_\mu^j = NER(\mathcal{H}_\mu^j) 
\end{equation}

To avoid semantic looping and encourage the discovery of new evidence, we only retain entities that have not yet been seen:

\begin{equation}
\label{eq:N_j_segnato}
\overline{\cal N}_\mu^j = {\cal N}_\mu^j \setminus \bigcup_{r=1}^{j-1} {\cal N}_\mu^r
\end{equation}

If $\overline{\cal N}_\mu^j = \emptyset$, the retrieval process terminates. Otherwise, we update the cumulative entity set:

\begin{equation}
\label{eq:N_j_total}
{\cal N}^j_{\mu_{total}} = \bigcup_{r=1}^{j} \overline{\cal N}^r_\mu
\end{equation}

We assume that ${\cal N}^j_{\mu_{total}}$ consists of $z$ entities:

\begin{equation}
\label{eq:N_j_total_1}
{\cal N}^j_{\mu_{total}}= \{ e_1, e_2, \cdots, e_z \}
\end{equation}

Let $t^j_{\mu_{total}} = \bigoplus_{x=1..z} e_x$ denote the textual concatenation of all entities in ${\cal N}^j_{\mu_{total}}$. We then construct the expanded query:

\begin{equation}
\label{eq:mu_j_exp}
\mu^j_{exp} = \mu \oplus t^j_{\mu_{total}}   
\end{equation}

together with an expanded set of embeddings:

\begin{equation}
\label{eq:E_j_exp}
{\cal E}^j_{\mu_{exp}} = \{ {\cal M}(\mu \oplus e_x) | e_x \in {\cal N}^j_{\mu_{total}} \}
\end{equation}

We use the expanded query $\mu^j_{exp}$ to reapply the same graph-based retrieval and candidate selection procedure described previously. Instead, the neural reranking stage works on the expanded embedding set ${\cal E}^j_{\mu_{exp}}$. This design maintains the semantic connection to the original query throughout the multi-hop exploration process and prevents the expanded entity context from overriding the retrieval objective. The resulting ranked chunks define the next retrieval hop $\mathcal{H}_\mu^{j+1}$.

Finally, we merge the retrieved candidate sets from all executed hops. In particular, let $h$ be the number of executed hops. For any chunk $c$ detected during the retrieval hops, we retain its maximum relevance score across all hops:

\begin{equation}
\label{eq:sigma_final}
\sigma_{c \mu}^{final} = \max_{u=1..h} \sigma^u_{c \mu}
\end{equation}

After computing the final relevance scores for all chunks, we construct a list ${\cal L}_\mu$ of chunks sorted by their corresponding final scores in a descending order. Finally, we construct the list ${\cal H}_\mu^{final}$ by selecting the top $\beta$ chunks from ${\cal L}_\mu$. ${\cal H}_\mu^{final}$ represents the final context window.

\section{Experimental Campaign}
\label{sec:Experimental-Campaign}

In this section, we describe the experimental campaign that we conducted to validate TIGRAG. Specifically, Subsection \ref{sub:Experimental-Setup} presents the experimental setup, and Subsection \ref{sub:Results} illustrates the obtained results. 

\subsection{Experimental Setup}
\label{sub:Experimental-Setup}

In this section, we describe the experimental setup for our tests. Specifically, we outline the datasets used, the baselines considered, the model and prompting protocol, the evaluated metrics, and finally the hyperparameter configuration. We conducted all experiments on a server with 256 AMD EPYC 7742 CPUs, 1 TB of RAM, and two NVIDIA A100 GPUs, each with 40 GB of memory. To ensure a fair comparison among the evaluated methods, we executed all indexing, retrieval, and generation experiments on the same hardware infrastructure.

\subsubsection{Datasets}
\label{subsub:Datasets}

We evaluated \name{}'s retrieval and reasoning capabilities across three widely adopted multi-hop Question Answering datasets: HotpotQA \cite{Yang*18}, 2WikiMultiHopQA \cite{Ho*20}, and MuSiQue \cite{Trivedi*22}. To mitigate the computational overhead while maintaining statistical significance, we subsampled $1,000$ queries from each dataset, following standard practices from previous studies \cite{Gutierrez*24, Trivedi*23, Press*23}. Each dataset provided questions to be answered and a corpus of associated documents. This corpus served as the evidence pool from which relevant context had to be retrieved. Furthermore, each query was paired with a ground-truth answer, enabling a rigorous assessment during the downstream evaluation phase. 

Table \ref{tab:statistics} provides a comprehensive overview of the statistics of the graph generated from each dataset.

\begin{table}[ht]
\centering
\scriptsize
\begin{tabular}{lrrr}
\toprule
{\bf Method} & {\bf HotpotQA} & {\bf 2WikiMultiHopQA} & {\bf MuSiQue} \\
\midrule
Number of nodes & 55,733 & 35,251 & 56,467 \\
Number of edges    & 1,315,330 & 658,326 & 1,370,777 \\
Average degree  & 47.20 & 37.35 & 48.55 \\
Average number of documents per node  & 8.15 & 6.99 & 8.18 \\
Number of chunks & 11,851 & 7,870 & 13,067 \\
Density & 0.0008 & 0.0011 & 0.0009 \\
Number of connected components & 1 & 1 & 1 \\

\bottomrule
\end{tabular}
\caption{Graph statistics for each dataset}
\label{tab:statistics}
\end{table}

\subsubsection{Baselines}
\label{subsub:Baselines}

To rigorously evaluate the effectiveness of \name{}, we benchmarked it against a diverse set of state-of-the-art RAG methods, covering both conventional dense retrieval techniques and more advanced structure-aware architectures. Specifically, we used {\em NaiveRAG} as a standard dense retrieval baseline, and employed {\em RAPTOR} \cite{Sarthi*24}, {\em GraphRAG} \cite{Edge*24}, {\em LightRAG} \cite{Guo*25-2}, {\em ToG} \cite{Sun*24-2}, {\em KGP} \cite{Wang*24-5}, and {\em HippoRAG2} \cite{Gutierrez*25} as recent approaches incorporating hierarchical, graph-based, or knowledge-guided retrieval mechanisms.  We also considered the case in which no RAG is used. To ensure fairness, we used the same generation prompt, designed to strictly enforce evidence-grounded reasoning. It is shown in Figure \ref{fig:prompt-template}. 


\subsubsection{Models and prompting protocol}
\label{subsub:Model-Prompting-Protocol}

The LLM we used for our experiments was \texttt{gemma3:27b-it-qat}\footnote{\url{https://ollama.com/library/gemma3:27b-it-qat}}. During the generation phase, we supplied it with the user question and the retrieved context via a prompt template shown in Figure ~\ref{fig:prompt-template}. Finally, to ensure a fair comparison, we standardized the generation stage across all three evaluated baselines by applying an equivalent prompting strategy. We employed the Nomic Text 1.5 model \footnote{\url{https://ollama.com/library/nomic-embed-text:v1.5}} for dense retrieval and semantic similarity computations. We performed tokenization using NLTK, which is based on the Treebank tokenizer and Punkt sentence segmentation. We extracted named entities using the spacy NER pipeline.

\begin{figure}[ht!]
\centering
\fbox{
\begin{minipage}{0.95\linewidth}
\scriptsize\ttfamily
You are a strict information extraction assistant. \\
Examine the context to find the exact answer. \\
You MUST follow this format: \\
Thought: <your step-by-step reasoning> \\
Answer: <the concise answer only> \\
If the answer is in the text, you must find it. Do not chat.\\

Example: \\
Context: \\ Wikipedia Title: The Last Horse\\ The Last Horse (Spanish: El último caballo) is a 1950 Spanish comedy film directed by Edgar Neville starring Fernando Fernán Gómez. \\
Question: Who directed The Last Horse?\\
Thought: The text states that the film `The Last Horse' was directed by Edgar Neville. \\
Answer: Edgar Neville.\\
---\\ \\
Context: <CONTEXT\_STR>\\
Question: <QUESTION>\\
Thought: \\
Answer: \\
\end{minipage}}
\caption{Prompt template used during the generation phase for Question Answering. {\tt <CONTEXT\_STR>} represents the retrieved context, while {\tt <QUESTION>} is the question to be answered.}
\label{fig:prompt-template}
\end{figure}

\subsubsection{Metrics}
\label{subsub:Metrics}

Following the procedure of \cite{Gutierrez*24}, we evaluated TIGRAG across three dimensions: retrieval efficacy, downstream generation quality, and computational efficiency. 

We quantified retrieval performance using Recall@2 (R@2) and Recall@5 (R@5), which measure the proportion of queries for which gold-standard supporting evidence is successfully retrieved within the top two (R@2) or top five (R@5) candidate chunks. For the generative Question Answering phase, we assessed performance using Exact Match (EM) and token-level F1. EM evaluates strict string equivalence, while token-level F1 evaluates the overlap between the LLM's output and the ground-truth answers at the token level. All of these metrics fall within the real range $[0,1]$, or $[0,100]$ if we express them in percentages; the higher their value, the better the performance. We omitted Recall@$k$ for KGP, ToG, and RAPTOR because their structural retrieval mechanisms are incompatible with chunk-level evaluation. In fact, KGP and ToG retrieve KG entities and relationships, while RAPTOR retrieves hierarchical summary nodes. Thus, these three systems do not permit reliable evaluation of retrieval quality through Recall. Finally, we omitted Recall@k for NoRAG because this system has no retrieval mechanism. 

In addition to accuracy, we evaluated the computational load of RAG approaches. Specifically, we measured the time required for offline graph construction in order to evaluate scalability with large corpora. For the online phase, we reported end-to-end inference latency, capturing both retrieval and LLM generation. We also measured the prompt footprint as the average token length of the retrieved context. This computational metric is critical because reducing the context footprint not only lowers computational overhead and latency but also helps mitigate the loss of focus during generation.

\subsubsection{Hyperparameter configuration}
\label{subsub:Hyperparameter-Configuration}

We evaluated TIGRAG using a fixed hyperparameter configuration. In particular, we set the threshold $th_\rho$ to 0.8 (see Equation \ref{eq:alpha}), the hyperparameter $\iota$ (denoting the cardinality of ${\cal H}_\mu^j$ - see Section \ref{sub:Multi-Hop-Reasoning}) to $1$ and the number $h$ of hops to $2$ (see Section \ref{sub:Multi-Hop-Reasoning}).  Furthermore, we set the hyperparameter $\beta$ (denoting the number of chunks composing the final context window - see Section \ref{sub:Multi-Hop-Reasoning}) to 5. We also set the length $w$ of the token co-occurrence window to 4. Finally, we set the window size $s$ to 6 and the overlap $o$ to 2; therefore, the stride length $l$ was set to 4 (see Section \ref{sub:Preprocessing-Text-Chunking}). In Section~\ref{subsub:Hyperparameter-Ablation}, we report a hyperparameter and ablation study supporting this configuration.

\subsection{Results}
\label{sub:Results}

In this section, we present the results obtained from our experimental campaign. Specifically, Subsection \ref{subsub:Performance-Comparison} compares the performance of \name{} with that of other related systems proposed in the literature. Subsection \ref{subsub:Resource-Comparison} presents a comparison of the computational resources required by \name{} and other related approaches. Finally, Subsection \ref{subsub:Hyperparameter-Ablation} presents a hyperparameter and ablation study of \name{}.

\subsubsection{Performance Comparison}
\label{subsub:Performance-Comparison}

Table~\ref{tab:Comparison-Results-recall} reports the retrieval performance of the compared methods in terms of R@2 and R@5 across the considered multi-hop Question Answering benchmarks. For each column, the number in bold indicates the maximum value, while the underlined number denotes the sub-maximum value.

\begin{table}[ht]
\centering
\tiny
\begin{tabular}{lcccccc|cc}
\toprule
{\bf Method} & \multicolumn{2}{c}{\bf HotpotQA} & \multicolumn{2}{c}{\bf 2WikiMultiHopQA} & \multicolumn{2}{c}{\bf MuSiQue} & \multicolumn{2}{c}{\bf Average} \\
\cmidrule(r){2-3} \cmidrule(r){4-5} \cmidrule(r){6-7} \cmidrule(l){8-9}
 & {\bf R@2} & {\bf R@5} & {\bf R@2} & {\bf R@5} & {\bf R@2} & {\bf R@5} & {\bf R@2} & {\bf R@5} \\
\midrule
HippoRAG2 & 57.45\% & 84.05\% & 63.42\% & \underline{78.73\%} & \underline{35.50\%} & \underline{51.29\%} & 52.12\% & \underline{71.36\%} \\
LightRAG & 69.05\% & 85.25\% & 59.98\% & 69.58\% & 35.62\% & 49.68\% & 54.88\% & 68.17\% \\
NaiveRAG & \underline{71.65\%} & \underline{85.05\%} & \underline{64.35\%} & 70.40\% & 38.39\% & 51.21\% & \underline{58.13\%} & 68.89\% \\
\midrule
\name{} & \textbf{78.60\%} & \textbf{92.05\%} & \textbf{72.98\%} & \textbf{87.20\%} & \textbf{42.57\%} & \textbf{56.28\%} & \textbf{64.72\%} & \textbf{78.51\%} \\
\bottomrule
\end{tabular}
\caption{Retrieval performance (R@2 and R@5) on multi-hop Question Answering benchmarks}
\label{tab:Comparison-Results-recall}
\end{table}

This table shows that \name{} outperforms all baselines consistently across all evaluated scenarios. On HotpotQA, it improves R@2 by 9.70\% and R@5 by 8.23\% compared to NaiveRAG. This trend continues on 2WikiMultiHopQA, where \name{} achieves a 13.41\% gain on R@2 over NaiveRAG, and a 10.76\% gain on R@5 over HippoRAG2. On the more challenging MuSiQue benchmark, which requires multi-step semantic reasoning, TIGRAG's margin of improvement increases further, with a 19.92\% gain on R@2 and a 9.73\% gain on R@5 over HippoRAG2. These results demonstrate the effectiveness of token-level topological expansion in enhancing multi-hop retrieval.

Afterwards, we shifted our focus from retrieval metrics to downstream generative performance. Table~\ref{tab:Comparison-Results-emf1} shows the Question Answering performance of the compared methods on the multi-hop benchmarks using Exact Match (EM) and token-level F1 (F1).

\begin{table}[ht]
\centering
\tiny
\begin{tabular}{lrrrrrr|rr}
\toprule
{\bf Method} & \multicolumn{2}{c}{\bf HotpotQA} & \multicolumn{2}{c}{\bf 2WikiMultiHopQA} & \multicolumn{2}{c}{\bf MuSiQue} & \multicolumn{2}{c}{\bf Average} \\
\cmidrule(r){2-3} \cmidrule(r){4-5} \cmidrule(r){6-7} \cmidrule(l){8-9}
 & \multicolumn{1}{c}{\bf EM} & \multicolumn{1}{c}{\bf F1}
 & \multicolumn{1}{c}{\bf EM} & \multicolumn{1}{c}{\bf F1}
 & \multicolumn{1}{c}{\bf EM} & \multicolumn{1}{c}{\bf F1}
 & \multicolumn{1}{c}{\bf EM} & \multicolumn{1}{c}{\bf F1} \\
\midrule
NoRAG & 28.30\% & 37.93\% & 30.90\% & 35.20\% & 7.40\% & 17.59\% & 22.20\% & 30.24\% \\
HippoRAG2 & 52.80\% & 64.40\% & \underline{50.20\%} & \underline{55.57\%} & \textbf{26.50\%} & \underline{33.89\%} & \underline{43.17\%} & \underline{51.29\%} \\
Raptor & 39.80\% & 55.69\% & 32.40\% & 38.24\% & 14.40\% & 24.21\% & 28.87\% & 39.38\% \\
KGP & 22.50\% & 32.26\% & 11.40\% & 13.08\% & 11.10\% & 18.15\% & 15.00\% & 21.16\% \\
ToG & 20.60\% & 26.63\% & 14.50\% & 17.02\% & 5.80\% & 8.60\% & 13.63\% & 17.42\% \\
GraphRAG & 29.99\% & 39.91\% & 7.60\% & 8.41\% & 0.50\% & 1.67\% & 12.70\% & 16.66\% \\
LightRAG & 52.50\% & 66.91\% & 45.70\% & 51.17\% & 22.20\% & 32.14\% & 40.13\% & 50.07\% \\
NaiveRAG & \underline{53.40\%} & \underline{64.84\%} & 44.40\% & 48.49\% & 21.00\% & 29.88\% & 39.60\% & 47.74\% \\
\midrule
\name{} & \textbf{60.70\%} & \textbf{72.60\%} & \textbf{50.90\%} & \textbf{58.60\%} & \underline{26.20\%} & \textbf{35.80\%} & \textbf{45.93\%} & \textbf{55.67\%} \\
\bottomrule
\end{tabular}
\caption{Question answering performance (EM and F1) on multi-hop Question Answering benchmarks}
\label{tab:Comparison-Results-emf1}
\end{table}

As shown in Table \ref{tab:Comparison-Results-emf1}, \name{} improves answer quality on HotpotQA consistently, achieving gains of 13.67\% in EM and 11.97\% in F1 over NaiveRAG. A similar trend is observed on 2WikiMultiHopQA, where \name{} surpasses HippoRAG2 by 1.39\% in EM and 5.45\% in F1. On the MuSiQue benchmark, HippoRAG2 obtains a marginally higher EM value, but \name{} preserves a higher F1 value. Since EM heavily penalizes partially correct but more detailed answers, TIGRAG's stronger F1 results suggest that our framework provides more comprehensive and semantically reliable evidence for multi-hop reasoning.

\subsubsection{Resource Comparison}
\label{subsub:Resource-Comparison}

The practical viability of a RAG architecture depends not only on retrieval accuracy but also on computational efficiency. After analyzing retrieval performance in Section \ref{subsub:Performance-Comparison}, we now focus on resource consumption during the offline and online phases. To this end, we analyze three critical dimensions: offline graph construction time (also referred to as indexing time), online end-to-end inference latency, and prompt footprint (quantified by the retrieved context's length expressed in number of tokens). These metrics evaluate \name{}'s scalability and practical usability. Offline indexing time measures the cost of processing large corpora, while end-to-end inference latency captures the responsiveness of the entire retrieval-generation pipeline. Excessively large contexts can slow down generation and distract the LLM; thus, reducing the context footprint is essential for improving both efficiency and reasoning quality. Table~\ref{tab:time} reports the offline indexing time and the average online inference latency per query obtained by TIGRAG and related approaches when performing the tests described in Section \ref{subsub:Performance-Comparison}.

\begin{table}[ht]
\scriptsize
\centering
\resizebox{\textwidth}{!}{
\begin{tabular}{lrrrrrr}
\toprule
{\bf Method} & \multicolumn{2}{c}{\bf HotpotQA} & \multicolumn{2}{c}{\bf 2WikiMultiHopQA} & \multicolumn{2}{c}{\bf MuSiQue}  \\
\cmidrule(r){2-3} \cmidrule(r){4-5} \cmidrule(r){6-7} 
 & {\bf Indexing (s)} & {\bf Retrieval (s)} & {\bf Indexing (s)}  & {\bf Retrieval (s)}  & {\bf Indexing (s)}  & {\bf Retrieval (s)}  \\
\midrule
HippoRAG2 & 308,664 & 75.60 & 121,945 & 79.48 & 318,282 & 81.67 \\
Raptor & 29,452 & 14.80 & 17,001 & 14.56 & 35,851 & 15.46    \\
KGP & 1,893 & 30.57 & 1,514 & 25.90 & 4,096 & 37.16 \\
ToG & 139,372 & 55.27 & 70,660 & 39.91 & 146,146 & 48.58 \\
GraphRAG & 657,055 & 14.91 & 74,802 & 15.75 & 195,325  &  15.75 \\
LightRAG & 721,833  & 20.31 & 343,812 & 19.96 & 593,942 & 20.46  \\
NaiveRAG & {\bf 102} & {\bf 3.35} & {\bf 66} & \underline{3.71} & {\bf 114} & \underline{4.45}  \\
\midrule
\name{}  & \underline{191} & \underline{3.72} & \underline{155} & {\bf 3.26} & \underline{202} &  {\bf 4.14}   \\
\bottomrule
\end{tabular}}
\caption{Indexing and retrieval time; the latter includes both the effective retrieval time and the LLM response time}
\label{tab:time}
\end{table} 

The analysis of this table shows that NaiveRAG has the shortest indexing time in the offline phase due to its simple, dense-retrieval architecture, which bypasses graph construction. Although \name{} is slower than NaiveRAG, it is still substantially faster than other graph-based methods while achieving stronger retrieval and generation performance (see Table \ref{tab:Comparison-Results-recall}).

A compelling trend emerges during the online inference phase. On HotpotQA, \name{} is slightly slower than NaiveRAG. However, on more complex datasets, such as 2WikiMultiHopQA and MuSiQue, \name{} surpasses NaiveRAG in end-to-end speed. This efficiency gain validates our hypothesis that, by dynamically extracting highly concentrated and relevant context, \name{} can drastically reduce the LLM's prompt processing and generation time. This reduction more than offsets the minimal latency introduced by the graph traversal necessary for PPR. Compared to RAPTOR, the fastest graph-based baseline, \name{} is roughly four times faster, setting a highly scalable standard for graph-augmented RAG systems.

Table~\ref{tab:contextlen} details the average context footprint, measured in number of tokens, generated per query. 

\begin{table}[ht]
\scriptsize
\centering
\begin{tabular}{lrrr}
\toprule
{\bf Method} & {\bf HotpotQA} & {\bf 2WikiMultiHopQA} & {\bf MuSiQue} \\
\midrule

HippoRAG2 & 579.45 & 639.43 & 656.61 \\
Raptor    & 867.88 & 860.08 & 869.98 \\
GraphRAG  & 3,843.83 & 888.46 & 1421.63 \\
KGP & 2,501.16 & \underline{367.28} & 2,323.68 \\
ToG & {\bf 117.96} & {\bf 117.10} & {\bf 115.30} \\
LightRAG  & 5,893.51 & 5,579.22 & 7,304.93 \\
NaiveRAG  & 742.80 & 719.44 & 800.26 \\
\midrule
\name{}   & \underline{578.02} & 570.66 & \textbf{635.41} \\
\bottomrule
\end{tabular}
\caption{Average context length (in number of tokens) across datasets}
\label{tab:contextlen}
\end{table}

As this table shows, approaches that rely strictly on dynamically filtered text chunks produce a significantly more compact context. In contrast, frameworks that augment their prompts by injecting extensive sets of KG triplets or community summaries, such as GraphRAG and LightRAG, suffer from highly inflated context windows. TIGRAG reduces prompt size by relying on a compact and highly relevant chunk pool, thereby improving inference speed and limiting the loss of focus.

\subsubsection{Hyperparameter and Ablation Study}
\label{subsub:Hyperparameter-Ablation}

First, we examine how the retrieval budget affects the way an LLM constructs context, with a specific focus on the number $\beta$ of chunks supplied to the LLM for context construction (see Section \ref{sub:Multi-Hop-Reasoning}). By varying $\beta$, we evaluate the tradeoff between providing sufficient supporting evidence and introducing semantic noise. Thus, we assess how context size ultimately influences EM and F1 scores. In Table~\ref{tab:Hyperparameter-c}, we present the results obtained from this experiment. In each column, we show in bold the best value obtained for the corresponding metric. The results of this table highlight that performance peaks at an intermediate value of $\beta$ ($\beta = 5$). Increasing the context window beyond this threshold decreases both EM and F1 scores. This behavior underscores LLMs' susceptibility to distraction when exposed to overly long and noisy contexts. Conversely, setting $\beta$ too low starves the generative model of necessary supporting evidence. Thus, $\beta = 5$ is the optimal configuration, which maximizes informational density without overwhelming the reasoning model's capacity.

\begin{table}[!ht]
\scriptsize
\centering
\begin{tabular}{lrrrrrr}
\toprule
 & \multicolumn{2}{c}{\bf HotpotQA} & \multicolumn{2}{c}{\bf 2WikiMultiHopQA} & \multicolumn{2}{c}{\bf MuSiQue}  \\
\cmidrule(r){2-3} \cmidrule(r){4-5} \cmidrule(r){6-7} 
$\mathbf{\beta}$ & {\bf EM} & {\bf F1} & {\bf EM} & {\bf F1} & {\bf EM} & {\bf F1} \\
\midrule
3 &  57.40\%  &  70.02\%  &  45.20\%  &  52.41\%  &  23.30\%  &  32.68\%  \\
5 &   \textbf{60.70\%}  &  \textbf{72.60\%}   &   \textbf{50.90\%}  &  \textbf{58.60\%}   &   \textbf{26.20\%}  &   \textbf{35.80\%}   \\
10 &  55.80\%  &  69.21\%  &  44.90\%  &  52.66\%  &  22.80\%  &   33.09\%   \\
\bottomrule
\end{tabular}
\caption{Sensitivity analysis for the hyperparameter $\beta$}
\label{tab:Hyperparameter-c}
\end{table}

Table~\ref{tab:Hyperparameter-h} reports the impact of the hyperparameter $\iota$, which denotes the cardinality of ${\cal H}_\mu^j$ (see Section \ref{sub:Multi-Hop-Reasoning}) on the effectiveness of retrieval, measured in terms of R@2 and R@5. In this table, we note that $\iota$ directly influences the precision of the retrieval space. Empirical results indicate that the optimal configuration is achieved with $\iota=1$ across most scenarios. This suggests that the essential bridging context is concentrated in the highest-ranked chunk, and incorporating subsequent chunks introduces distracting entities. In fact, in the latter case, the entity pool used for query augmentation excessively expands, inevitably introducing semantic drift and degrading overall performance. The only exception is observed on the 2WikiMultiHopQA dataset, where R@5 peaks at $\iota=2$.

\begin{table}[!ht]
\scriptsize
\centering
\begin{tabular}{lrrrrrr}
\toprule
$\iota$ & \multicolumn{2}{c}{\bf HotpotQA} & \multicolumn{2}{c}{\bf 2WikiMultiHopQA} & \multicolumn{2}{c}{\bf MuSiQue} \\
& {\bf R@2} & {\bf R@5} & {\bf R@2} & {\bf R@5} & {\bf R@2} & {\bf R@5} \\ \midrule
1 &  \textbf{78.60\%}  &  \textbf{92.05\%}  & \textbf{72.98\%} & 87.20\% &  \textbf{42.57\%}  &  \textbf{56.28\%}  \\
2 &  74.05\%  &  91.45\% &  63.52\%  &  \textbf{90.58\%}  &  37.78\%  &  55.17\%   \\
3 &  68.00\%  &  91.10\%  &  55.13\%  &  88.18\% &  33.92\%  &  55.16\%    \\
\bottomrule
\end{tabular}
\caption{Sensitivity analysis for the hyperparameter $\iota$}
\label{tab:Hyperparameter-h}
\end{table}

Table~\ref{tab:Hyperparameter-hops} shows how the hyperparameter $h$, which denotes the number of query hops (see Section \ref{sub:Multi-Hop-Reasoning}), affects the \name{}'s retrieval performance. We carried out this evaluation considering one, two, and three hops, aligning this hyperparameter with the multi-hop complexity of the underlying benchmarks. We did not explore beyond three hops because that exceeds the maximum ground-truth reasoning chain required by the selected datasets. Thus, additional hops would primarily introduce computational overhead and semantic noise rather than useful evidence. As shown in this table, using two hops yields the best tradeoff between semantic expansion and topological noise across all benchmarks.

\begin{table}[!ht]
\scriptsize
\centering
\begin{tabular}{lrrrrrr}
\toprule
$h$ & \multicolumn{2}{c}{\bf HotpotQA} & \multicolumn{2}{c}{\bf 2WikiMultiHopQA} & \multicolumn{2}{c}{\bf MuSiQue} \\
& {\bf R@2} & {\bf R@5} & {\bf R@2} & {\bf R@5} & {\bf R@2} & {\bf R@5} \\ \midrule
1 &  71.50\%  &  85.55\%  &  61.92\%  &  70.27\%  &  38.01\%  &  50.72\%  \\
2 &  \textbf{78.60\%}  &  \textbf{92.05\%}  & \textbf{72.98\%} & \textbf{87.20\%} &  \textbf{42.57\%}  &  \textbf{56.28\%}  \\
3 &  78.40\%  &  91.90\%  &  72.75\%  &  87.18\%  &  42.47\%  &  56.19\%   \\
\bottomrule
\end{tabular}
\caption{Sensitivity analysis for the hyperparameter $h$}
\label{tab:Hyperparameter-hops}
\end{table}

Table~\ref{tab:Hyperparameter-tau} evaluates the effect of the cumulative score threshold $th_\rho$ used in the candidate pooling stage (see Equation \ref{eq:alpha}). As shown in this table, higher values of $th_\rho$ improve retrieval effectiveness by enabling broader semantic exploration before the final reranking stage, filtering out the least relevant evidence. Performance consistently peaks when $th_\rho = 0.8$, suggesting that a large and diverse candidate pool benefits multi-hop reasoning. A lower value of $th_\rho$, instead, would introduce overly aggressive pruning and would limit access to interconnected evidence.

\begin{table}[!ht]
\scriptsize
\centering
\begin{tabular}{lrrrrrr}
\toprule
$th_\rho$ & \multicolumn{2}{c}{\bf HotpotQA} & \multicolumn{2}{c}{\bf 2WikiMultiHopQA} & \multicolumn{2}{c}{\bf MuSiQue} \\
& {\bf R@2} & {\bf R@5} & {\bf R@2} & {\bf R@5} & {\bf R@2} & {\bf R@5} \\ \midrule
0.1 &  78.30\%  & 91.70\% &  72.78\%  &  87.05\%  &  41.83\%  &  54.49\%  \\
0.2 &  78.55\%  & 92.05\%  &  72.72\%  &  86.92\%  &  42.05\%  &  55.09\%  \\
0.5 & \textbf{78.60\%} & \textbf{92.10\%} &  72.52\%  &  86.75\%  &  42.42\%  &  55.91\%  \\
0.8 &  \textbf{78.60\%}  &  92.05\%  & \textbf{72.98\%} & \textbf{87.20\%} &  \textbf{42.57\%}  &  \textbf{56.28\%}  \\
0.9 &  \textbf{78.60\%}  &  92.05\% &  72.92\%  &  87.15\%  &  42.57\%  &  56.28\%  \\
\bottomrule
\end{tabular}
\caption{Sensitivity analysis for the hyperparameter $th_\rho$}
\label{tab:Hyperparameter-tau}
\end{table}

Table~\ref{tab:Ablation} presents an ablation study designed to isolate the contributions of \name{}'s main components. Specifically, we evaluated: {\em (i)} a \name{} variant without BM25, where retrieval relies solely on graph topology-aware matching; {\em (ii)} a \name{} variant based on degree centrality instead of PPR-driven expansion; and {\em (iii)} a \name{} variant without the final neural reranking stage.

\begin{table}[!ht]
\tiny
\centering
\begin{tabular}{lcccccc}
\toprule
 {\bf Approach} & \multicolumn{2}{c}{\bf HotpotQA} & \multicolumn{2}{c}{\bf 2WikiMultiHopQA} & \multicolumn{2}{c}{\bf MuSiQue} \\
& {\bf R@2} & {\bf R@5} & {\bf R@2} & {\bf R@5} & {\bf R@2} & {\bf R@5} \\ \midrule
\name{} without BM25 & 78.55\%  & 92.00\% &  72.92\%  &  87.15\%  &  42.33\%  &  55.91\%  \\
\name{} based on degree centrality & 22.00\% &  24.15\%  & 44.97\% & 51.98\% & 12.86\%  &  16.54\%  \\
\name{} without reranking &  51.10\% & 69.00\% &  48.18\%  &  64.32\%  &  29.53\%  &  41.40\%  \\
\name{} & \textbf{78.60\%}  &  \textbf{92.05\%}  & \textbf{72.98\%} & \textbf{87.20\%} &  \textbf{42.57\%}  &  \textbf{56.28\%} \\
\bottomrule
\end{tabular}
\caption{Ablation study evaluating the contribution of the main components of \name{} on retrieval performance across the considered benchmarks}
\label{tab:Ablation}
\end{table}

The ablation results highlight the importance of each component of the proposed framework. Removing BM25 yields only marginal performance drops, confirming that the graph-based expansion mechanism alone captures highly relevant evidence. Replacing PPR with degree centrality dramatically decreases performance across all datasets, demonstrating the importance of topology-aware semantic propagation over simple node degree. Finally, eliminating the neural reranking stage significantly reduces retrieval effectiveness, indicating that the reranking is essential for filtering out noisy candidates introduced during graph expansion.

\section{Discussion}
\label{sec:Discussion}


From a performance point of view, the topological approach underlying \name{} enhances retrieval and consistently outperforms baselines. This high-fidelity retrieval directly translates into superior downstream Question Answering generation, setting new state-of-the-art Exact Match and token-level F1 score benchmarks. Beyond strict accuracy, the most significant implication of \name{} lies in its resource efficiency. By constructing the structural graph solely through statistical co-occurrences, \name{} eliminates the need for the computationally expensive LLM-driven entity extraction pipelines required by traditional GraphRAG approaches. Consequently, it consumes a fraction of the time and token budget during offline graph construction and online inference, offering a highly scalable solution for real-world deployment.

Despite these promising results, \name{} has some limitations that suggest interesting directions for future research. First,
\name{}'s iterative, entity-driven retrieval strategy relies heavily on the quality of the Named Entity Recognition (NER) pipeline. Since query expansion is guided by entities extracted from previously retrieved chunks, inaccurate, ambiguous, or irrelevant entities can spread across retrieval stages, resulting in semantic drift and reduced retrieval precision. Therefore, exploring more robust entity disambiguation and filtering mechanisms is an important direction for future work.
Second, as with other co-occurrence-based approaches, \name{} primarily captures relationships emerging from token proximity. Future work will investigate lightweight semantic enrichment strategies that can better model implicit connections while preserving TIGRAG's computational efficiency.

\section{Conclusion}
\label{sec:Conclusion}


In this work, we presented \name{}, an effective and highly efficient RAG framework driven by a token co-occurrence KG. Unlike traditional GraphRAG approaches, which rely on expensive LLM-based entity and relation extraction pipelines, \name{} constructs a lightweight topological representation directly from token co-occurrences. It also uses PPR-based semantic expansion and dynamic neural reranking to retrieve interconnected evidence across multiple hops. By bypassing computationally expensive LLM-based graph construction indexing, the token-level approach underlying \name{} enables faster offline graph construction while preserving the structural dependencies necessary for multi-hop reasoning. Empirical evaluations across HotpotQA, 2WikiMultiHopQA, and MuSiQue demonstrate that \name{} significantly improves retrieval performance, establishing new state-of-the-art performance compared to baseline models. These findings prove that graph-augmented RAG systems can solve complex, multi-hop reasoning tasks without sacrificing computational efficiency.

Our current evaluation focuses on textual, multi-hop datasets; however, this study suggests several promising directions for future research. First, we plan to expand the graph topology beyond statistical co-occurrences by introducing semantic edges. This semantic smoothing will connect lexically distinct but semantically similar tokens, capturing implicit relationships. Second, we plan to address the inherent limitations of treating tokens as only lexical units, by introducing semantic-aware nodes to address token polysemy. This will allow the graph to dynamically disambiguate meanings for more granular and precise retrieval. Finally, since modern information ecosystems are heterogeneous by nature, a future development is to extend \name{} to multimodal settings. This will involve integrating images and videos into the topological space to support more complex QA tasks.


\section*{Declaration of competing interest}
The authors declare that they have no known competing financial interests or personal relationships that could have appeared to influence the work reported in this paper.

\section*{Data availability}
The code used for our study will be available after publication. 

\bibliographystyle{plain}
\bibliography{bibliografia,temp}

\end{document}